\documentclass[letterpaper, 10 pt, conference]{ieeeconf}  

\IEEEoverridecommandlockouts                              
                                                          
\usepackage[utf8]{inputenc} 
\usepackage[T1]{fontenc}    
\usepackage{hyperref}       
\usepackage{url}            
\usepackage{booktabs}       
\usepackage{amsfonts}       
\usepackage{nicefrac}       
\usepackage{microtype}      
\usepackage{graphicx}
\usepackage{booktabs}   
\usepackage{xcolor}

\title{\LARGE \bf  A Systematic Search over Deep Convolutional Neural Network Architectures for Screening Chest Radiographs  }

\author{Arka Mitra$^{1}$, Arunava Chakravarty$^{1}$, Nirmalya Ghosh$^{1}$, Tandra Sarkar$^{2}$,\\ Ramanathan Sethuraman$^{3}$,  Debdoot Sheet$^{1}$
\thanks{*This work is supported through a research grant from Intel India Grand Challenge 2016 for Project MIRIAD.}
\thanks{$^{1}$ A. Mitra, A. Chakravarty, N. Ghosh, D. Sheet are with the Indian Institute of Technology Kharagpur, India-721302. A. Mitra and A. Chakravarty contributed equally to this work. {\tt \{arunava, nirmalya, debdoot\}@ee.iitkgp.ac.in}}%
\thanks{$^{2}$ T. Sarkar is with Apollo Gleneagles Hospital, Kolkata, India}%
\thanks{$^{3}$ R. Sethuraman is with Intel Technology India Pvt. Ltd. Bangalore, India }}

\begin{document}
\maketitle
\thispagestyle{empty}
\pagestyle{empty}

\begin{abstract}
Chest radiographs are primarily employed for the screening of pulmonary and cardio-/thoracic conditions. Being undertaken at primary healthcare centers, they require the presence of an on-premise reporting Radiologist, which is a challenge in low and middle income countries. This has inspired the development of machine learning based automation of the screening process. While recent efforts demonstrate a performance benchmark using an ensemble of deep convolutional neural networks (CNN), our systematic search over multiple standard CNN architectures identified single candidate CNN models whose classification performances were found to be at par with ensembles. Over $63$ experiments spanning $400$ hours, executed on a $11.3$ FP32 TensorTFLOPS compute system, we found the Xception and ResNet-18 architectures to be consistent performers in identifying co-existing disease conditions with an average AUC of 0.87 across nine pathologies. We conclude on the reliability of the models by assessing their saliency maps generated using the randomized input sampling for explanation (RISE) method and qualitatively validating them against manual annotations locally sourced from an experienced Radiologist. We also draw a critical note on the limitations of the publicly available CheXpert dataset primarily on account of disparity in class distribution in training vs. testing sets, and unavailability of sufficient samples for few classes, which hampers quantitative reporting due to sample insufficiency. 
\end{abstract}
\begin{keywords}
Chest X-ray, CNN, transfer learning
\end{keywords}
\section{Introduction}
Chest radiography is an inexpensive imaging modality that is routinely employed for the screening of several cardio-thoracic, pulmonary conditions and pathologies. Due to the shortage of Radiologists, currently there are delays in the reporting of screening tests~\cite{bastawrous2017improving, rimmer2017radiologist} which prevents timely intervention. The development of automated tools for the assesment of chest radiographs can help minimize the intra and inter-observer variation in reporting and reduce the time and effort of Radiologists and thereby enable them to focus on critical cases. 

Existing methods have explored handcrafted  multi-scale shape, edge and texture features for screening Tuberculosis in chest radiographs \cite{santosh2017automated}. Convolutional neural networks (CNN) have led to a significant improvement in performance with a majority of the existing methods \cite{chexpert}, \cite{8719904}, \cite{yang2019learn} adapting the DenseNet \cite{densenet} architecture to simultaneously detect the presence of multiple diseases in chest radiographs.  An ensemble  of handcrafted and deep learning based features extracted using the pre-trained VGG-16 and 19 architecures was employed in \cite{8856715} for Pneumonia detection, while \cite{baltruschat2019comparison} employed the ResNet-50 architecture with different resolutions of the X-ray images and integrated non-image data such as patient age, gender and acquisition type to improve the classification performance.

The lack of sufficient training data especially relevant to the domain of medical image analysis often inhibits the training of a CNN from scratch due to over-fitting. This has led to the popularity of transfer learning techniques for adapting the pre-trained weights of existing CNNs to solve a target task. Often, multiple architectures need to be assessed under different transfer learning protocols to find the optimal setting. In this context, this work explores possible solutions for screening chest radiographs for 14 disease classes through a systematic evaluation of a set of candidate CNN models by adapting them to chest X-ray images using different transfer learning protocols and ways to handle the uncertain labels in the training set. Our methodology is detailed in Section \ref{Sec:method}. The performance of the CNNs are quantitatively analyzed and validated qualitatively against the local annotations provided by an experienced Radiologist in Section \ref{Sec:result}. Additionally, the impact of employing an ensemble of CNN architectures as opposed to a single CNN model has also been evaluated.

\section{Method}
\label{Sec:method}

\begin{figure*}[t]
\centering
	\includegraphics[width=0.85\textwidth]{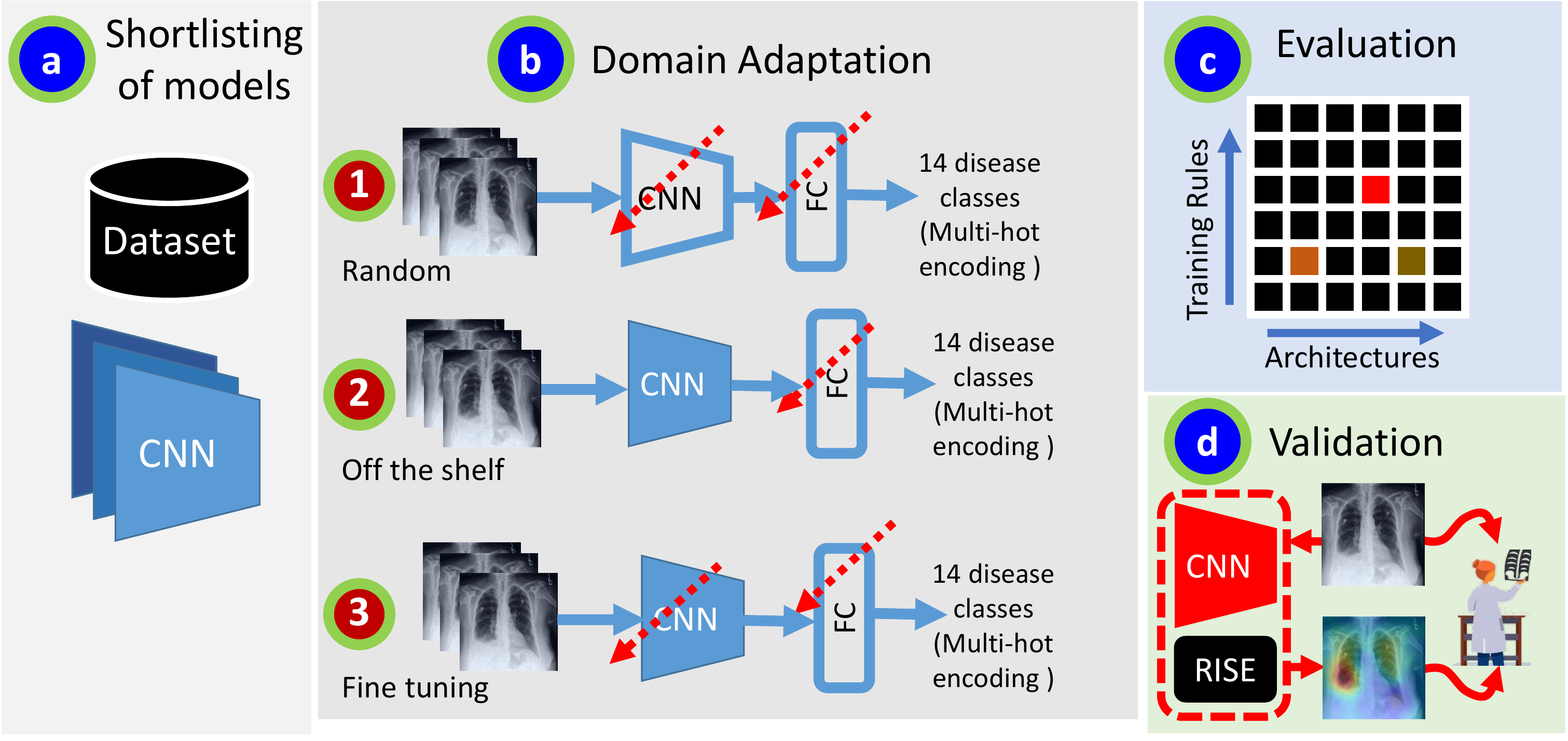}
	\caption{Overview of our approach for Radiologist validated systematic search over CNN architectures for chest radiograph screening. We start with (a) shortlisting a set of candidate models viz. VGG, ResNet, DenseNet, Inception, Xception; (b) adapting them by (1) either training the model end-to-end initialized at random weights; or using pre-trained weights (2) without updating them or (3) fine-tuning them; (c) quantitatively evaluating performance; (d) validating model reliability through a Radiologist's assessment of visual explainers.}
	\label{fig:ga}
\end{figure*}

The method is described in Fig. \ref{fig:ga}.  The \textbf{CNN architectures} used are VGG-16/19 with batch normalization (BN)~\cite{vgg}, GoogLeNet Inception-v3~\cite{inception}, ResNet-18/50~\cite{resnet}, DenseNet-121~\cite{densenet} and Xception~\cite{xception}.  The final classification layers for each CNN model is replaced by a Global Average Pooling to map the feature channels into a 1D vector. This is followed by a Fully Connected (FC) layer comprising 14 neurons with sigmoid activations (for mult-hot encoding) to detect the presence of one or more disease classes in each image.

The  \textbf{Image pre-processing} comprises resizing the 2D grayscale radiographs to $320 \times 320$, followed by replicating it to obtain 3 channel inputs for the CNNs. The three channels are normalized to match the statistics of the ImageNet~\cite{imagenet} dataset. Data augmentation is applied to the training images on-the-fly and consists of random horizontal flips and  random crops followed by resize operation to $320 \times 320$. 

\textbf{Adaptation} of the CNNs is performed in 3 ways: (1) Train the entire CNN from scratch after initialization with random weights (R); (2) The CNN weights are initialized with an off-the-shelf (O) ImageNet~\cite{imagenet} pre-trained model and not updated during learning while FC is randomly initialized and updated; (3) The CNN weights are initialized with ImageNet~\cite{imagenet} pre-trained model, the FC is initialized using the weights learned in setting (O) and the entire network is fine-tuned (F) in an end-to-end manner. 

The training dataset contains noisy Ground Truth (GT) labels for each disease with 0 indicating the absence, 1 indicating the presence and -1 indicating an uncertain label. Three settings were explored to \textbf{handle the uncertain labels}  by (i) replacing -1 with 0 (U-Zeros), ie, treating them as absence of the disease; (ii) replacing -1 with 1 (U-Ones), ie., treating them as presence of the disease and (iii) masking them out to have zero loss during training (U-Ignore).

The \textbf{Implementation details} are as follows. Experiments spanning $400$ hours are performed over a total of $63$ combinations obtained across $7$ CNN architectures $\times 3$ adaptation rules $\times 3$ ways to handle uncertain labels. 

Each CNN model is implemented in Anaconda Python 3, PyTorch 1.0 and trained  for $6$ epochs, $13,963$ batch updates per epoch with a batch size of $16$ using Adam~\cite{adam} optimizer, learning rate of $10^{-4}$, $\beta_1=0.9$, $\beta_2=0.999$ and a weight decay of $1\times 10^{-5}$.  The binary cross-entropy loss was employed for training \textit{without} a $Softmax(\cdot)$ operation  as multiple diseases could co-occur. 

The models are trained on a Ubuntu 16.04 LTS server with $2\times$ Intel Xeon 4110 CPU, $12\times 8$ GB DDR4 RAM, $2\times 2$ TB HDD, $4\times$ Nvidia GTX 1080Ti GPU with $11$ GB memory.

\section{Experiments}
\label{Sec:result}

\begin{figure*}[]
 \centering
  \includegraphics[scale=.32]{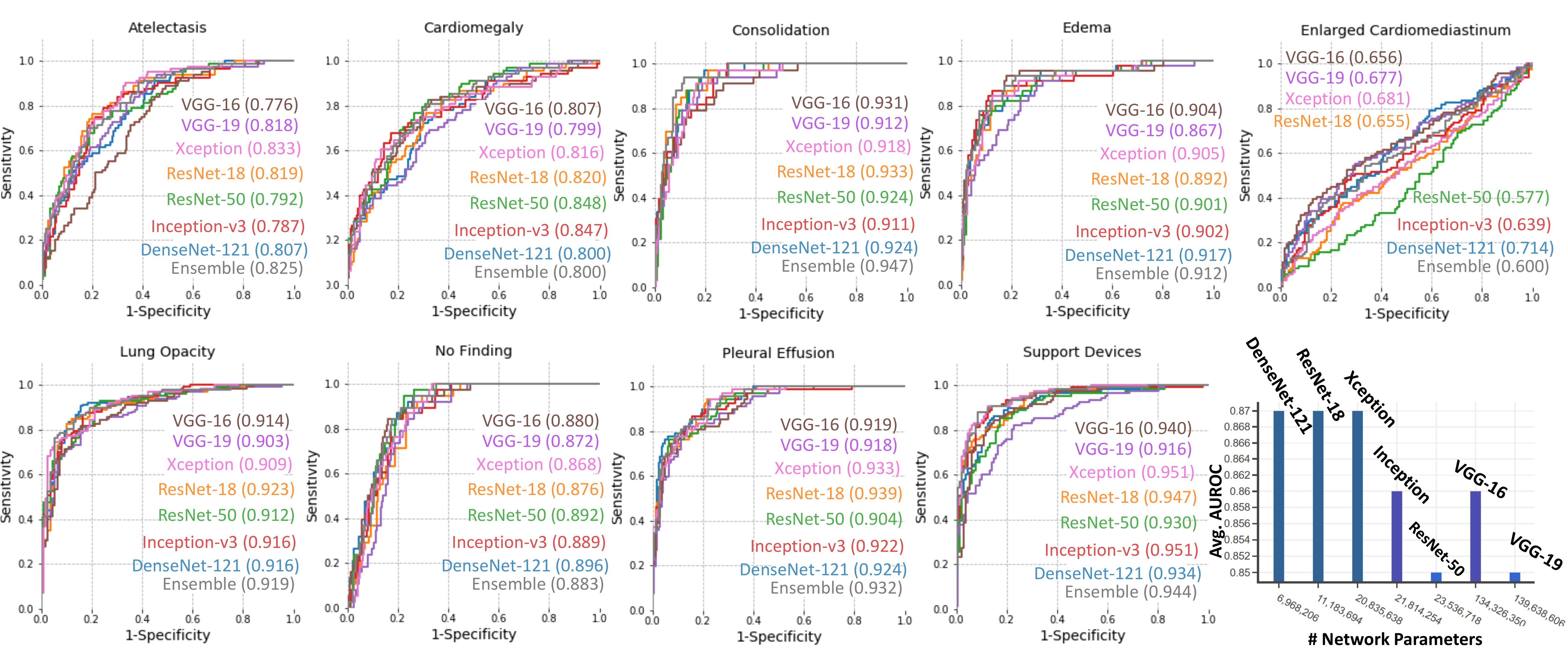}
 \caption{The ROC curves for the classification of nine diseases in Chest radiographs. The AUC values are provided in corresponding plot legend. The average AUC values across the 9 disease classes and the number of network parameters for each CNN architecture is presented in the bottom- right figure.}
\label{Fig:chexpert_roc} 
 \end{figure*}

\renewcommand{\arraystretch}{1}
\begin{table*}[]
\centering
\label{Tab:u_zero_small}
\caption{Area under the ROC curves (AUC) for the chest X-ray disease classification using  U-Zeros setting and reported are best performing models. The average AUC is computed across the nine disease classes with sufficient samples.}
\resizebox{\textwidth}{!}{
\begin{tabular}{@{}lccccccccccccc|c@{}}
\toprule
                 & Atelectasis & \begin{tabular}[c]{@{}l@{}}Cardio-\\ megaly\end{tabular} & Edema  & \begin{tabular}[c]{@{}l@{}}Consolid\\ -ation\end{tabular} & \begin{tabular}[c]{@{}l@{}}Pleural\\ Effusion\end{tabular} & \begin{tabular}[c]{@{}l@{}}Support\\ Devices\end{tabular} & \begin{tabular}[c]{@{}l@{}}Lung\\ Opacity\end{tabular} & \begin{tabular}[c]{@{}l@{}}Enlarged\\  Cardiom.\end{tabular} & \begin{tabular}[c]{@{}l@{}}No\\ Finding\end{tabular} & \begin{tabular}[c]{@{}l@{}}Pneum-\\ onia\end{tabular} & \begin{tabular}[c]{@{}l@{}}Pneumo-\\ thorax\end{tabular} & \begin{tabular}[c]{@{}l@{}}Lung\\ Lesion\end{tabular} & \begin{tabular}[c]{@{}l@{}}Pleural \\ Other\end{tabular} & Avg.\\ \midrule
                 
\# Training & 33376 & 27000   & 52246  & 14783   & 86187  & 116001 & 105581   & 10798  & 22381 & 6039   & 19448  & 9186  & 3523   \\ 
\# Testing & 80 & 68 & 45 & 33 & 67 & 107   & 126 & 68 & 38 & 8  & 8 & 1 & 1  \\ \midrule

Xception (F) & 0.833  & 0.816   & 0.905  & 0.918   & 0.933     & 0.951   & 0.909  & 0.681     & 0.868 & 0.684 & 0.757 & 0.923 & 0.974 & 0.87  \\ \midrule
DenseNet-121(F)& 0.807 & 0.800 & 0.917  & 0.924  & 0.924 & 0.934   & 0.916  & 0.714   & 0.896 & 0.723   & 0.777  & 0.584    & 0.918 & 0.87\\ \midrule
ResNet-18 (F) & 0.819  & 0.820 & 0.892  & 0.933  & 0.939   & 0.947 & 0.923   & 0.655  & 0.876 & 0.715 & 0.838   & 0.305 & 0.983 & 0.87\\ \midrule

Ensemble (Avg.)  & 0.8254 & 0.8002 & 0.9121 & 0.9465   & 0.9316 & 0.9436  & 0.9194  & 0.5955 & 0.8827 & 0.6455 & 0.8008 & 0.2489  & 0.8584 & 0.86\\ 
Ensemble (Wted.) & 0.8262  & 0.8004 & 0.9122 & 0.9466 & 0.9318 & 0.9437 & 0.9197 & 0.6031  & 0.8829 & 0.6665 & 0.8070 & 0.4592 & 0.8841  & 0.86\\ \midrule

Prior art \cite{chexpert} & 0.811 & 0.840 & 0.929  & 0.932 & 0.931   & --  & --  & --  & -- & -- & -- & -- & -- \\ \bottomrule
\end{tabular}
}
\end{table*}

\begin{figure*}[]
 \centering
  \includegraphics[width=0.9\textwidth]{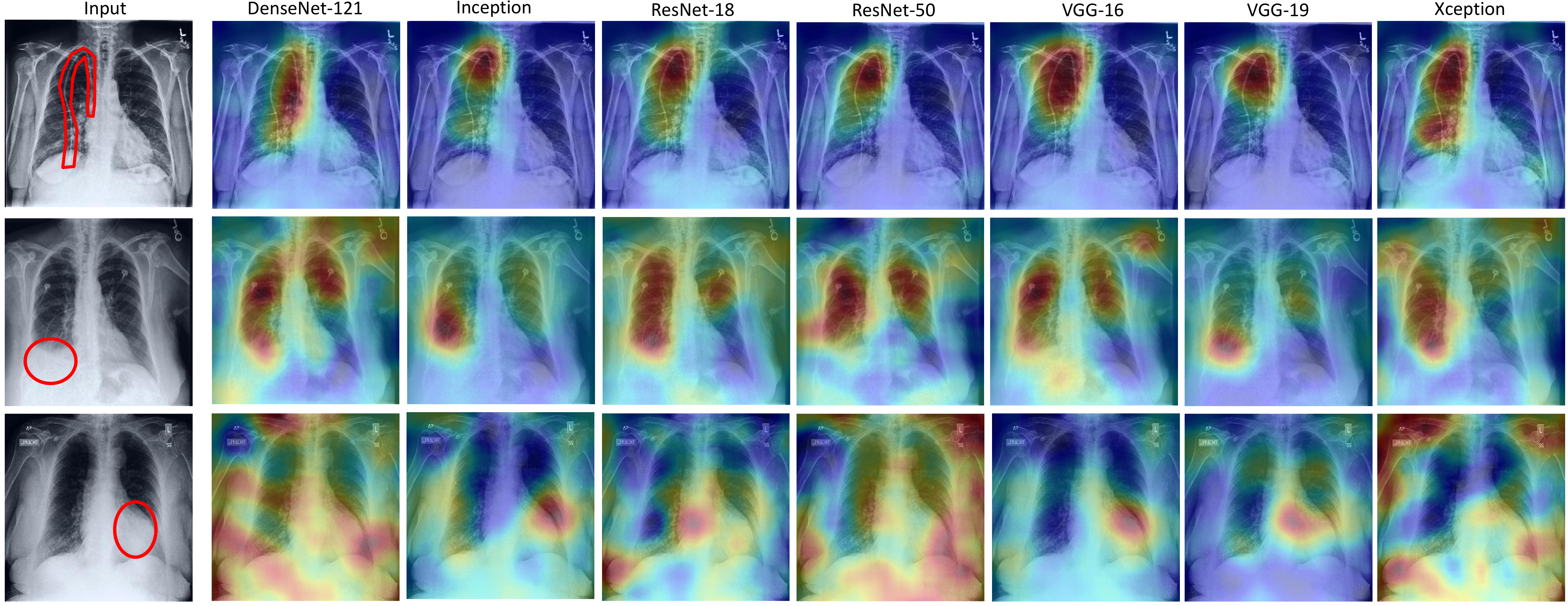}
 \caption{RISE \cite{rise} based saliency map visualization for each CNN. The abnormal region is marked in \textsc{Red} by a Radiologist. Top row: Supporting Device; Middle row: Pleural Effusion; Bottom row: Cardiomegaly.}
\label{Fig:chexpert_rise1} 
 \end{figure*}

\textbf{Dataset:} The CNN architectures have been evaluated on the Chexpert \cite{chexpert} dataset which  consists of $223,414$ training and $234$ test images with ground truth (GT) labels indicating the presence or absence of fourteen disease classes.   The images were acquired at varying resolutions, views and image quality.  The GT for training set is noisy and labeled as either present (1), absent(0) or uncertain (-1) as they were obtained from free-text radiology reports using an automated natural language processing (NLP) tool.  The GT for the test set did not have uncertain labels as they were obtained from the majority consensus opinion of 3 Radiologists. Multiple diseases can co-occur in the same image and the class-wise distribution of samples is  presented in the first two rows in Table I .

\textbf{Results and Discussion: }
Due to space limitations, the detailed performance of all 7 CNN architectures using different ways to handle the uncertain labels and transfer learning protocols during training is available online at \url{http://bit.do/Suppl_CXR_EMBC} (Tables II-IV). The metrics for the  three \textit{best performing models} are summarized in Table 1 and compared to the prior art and two ensemble models.
There were no test samples with fracture and less than 9 cases with pneumonia, pneumothorax, lung lesion and pleural other disease classes in the test set, implying that the results reported for these may not be representative of the actual performance. The average Area under the ROC curve (AUC) across the remaining nine classes is reported in the last column of Table I. The ROC plots and the corresponding AUC values of the best performing model across the different training settings
for each of the 7 CNN architectures is presented in Fig. 2.
 
 Fine-tuning(F) the entire CNN architecture consistently performed the best, while employing CNNs with pre-trained weights as off-the-shelf feature extractors performed the worst for all experiments (see Suppl. Tabs. II-IV). Among the different ways to handle the uncertain labels in the training set during Fine-tuning,  U-Zeros was found to perform well while U-Ignore performed the worst for most of the architectures except ResNet-50. All CNN architectures had an average AUC in the range [0.85, 0.87] using Fine-tuning (F) with ResNet-18, DenseNet-121 and Xception performing marginally better. In comparison to the performance of the five disease classes ( first five columns in Table I) reported in \cite{chexpert}, we found ResNet-18 alone to perform better than \cite{chexpert} on three out of the five classes with the exception of Cardiomegaly and Edema. 

A qualitative evaluation of the region where each CNN attended for the classification was performed by computing the saliency maps for each of the trained models using the Randomized Input Sampling for Evaluation (RISE) ~\cite{rise} method. The saliency maps were qualitatively compared with locally sourced annotations from a Radiologist on a subset of test images. Overall, the  Xception and ResNet-18 architectures seemed to show better correlation with the manual annotations. Few examples of different disease cases are depicted in Fig.~\ref{Fig:chexpert_rise1}.

In addition to the single CNN models, ensembles of the seven CNN architectures were also evaluated. The best performing model weight for each architecture across the different training protocols was considered for the ensemble construction. The ensemble decision was computed in two ways: a) an \textit{unbiased} ensemble decision was obtained by averaging the prediction scores from each CNN; b) a \textit{biased} ensemble decision was obtained by taking the \textit{weighted} average of the prediction scores from each CNN model for each disease class. For each CNN, different class-specific weights were employed which were proportional to the validation AUC scores of the CNN for a particular class. The weighted ensemble was found to have almost similar performance in comparison to the unbiased average ensemble except for the classes with very few test samples. Moreover, the ensemble methods were actually found to lead to the suppression of the individually strong models and didnot lead to any significant improvement in performance over the single best performing CNN models.

\section{Conclusion}
Screening of chest radiographs plays a crucial role in the diagnosis and management of several cardio, thoracic and pulmonary pathologies. Although CNNs have led to significant improvements in performance on image classification tasks, often a multitude of CNN architectures need to be assessed for a given task. In this work we explore possible solutions for screening chest radiographs for 14 disease classes through a systematic evaluation of a set of candidate CNN models by adapting them to chest X-ray images using different adaptation protocols and ways to handle uncertain labels in the training set. A comprehensive set of experiments was performed involving 7 CNN architectures, 3 ways to adapt the model weights and 3 ways to handle the uncertain labels. Limitations of the dataset primarily on account of disparity in class distribution in training vs. testing sets, and unavailability of sufficient samples for five out of the 14 disease classes hampers quantitative reporting due to sample insufficiency. Both quantitative performance on the remaining nine disease classes as well as the qualitative evaluation of the saliency maps for each CNN model by a Radiologist indicate Xception and ResNet-18 architectures to be most suitable for chest radiography screening with an average AUC of 0.87 while ensembles constructed using multiple CNN models did not show any significant performance improvement. Further refinement of the ResNet-18 and Xception architectures through EfficientNet~\cite{tan2019efficientnet} based search and improving their saliency map based explainability using an active learning based approach with a radiologist in the loop are promising directions for future work.

\bibliographystyle{unsrt}
\bibliography{root}


\onecolumn
\setcounter{page}{1}
\section*{Supplementary Materials}

\renewcommand{\arraystretch}{1}
\begin{table*}[!tbh]
\centering
\label{Tab:u_zero}
\caption{Area under the ROC curves (AUC) for the Chest X-ray disease classification using  U-Zeros setting to handle uncertain labels in the training set. The Average AUC is computed across the disease classes with sufficient samples in columns 1-9. The best performance of each architecture is indicated in bold for each disease.}
\resizebox{\textwidth}{!}{
\begin{tabular}{@{}lccccccccccccc|c@{}}
\toprule
                 & Atelectasis & \begin{tabular}[c]{@{}l@{}}Cardio-\\ megaly\end{tabular} & Edema  & \begin{tabular}[c]{@{}l@{}}Consolid\\ -ation\end{tabular} & \begin{tabular}[c]{@{}l@{}}Pleural\\ Effusion\end{tabular} & \begin{tabular}[c]{@{}l@{}}Support\\ Devices\end{tabular} & \begin{tabular}[c]{@{}l@{}}Lung\\ Opacity\end{tabular} & \begin{tabular}[c]{@{}l@{}}Enlarged\\  Cardiom.\end{tabular} & \begin{tabular}[c]{@{}l@{}}No\\ Finding\end{tabular} & \begin{tabular}[c]{@{}l@{}}Pneum-\\ onia\end{tabular} & \begin{tabular}[c]{@{}l@{}}Pneumo-\\ thorax\end{tabular} & \begin{tabular}[c]{@{}l@{}}Lung\\ Lesion\end{tabular} & \begin{tabular}[c]{@{}l@{}}Pleural \\ Other\end{tabular} & Avg.\\ \midrule
                 
\# Training & 33376 & 27000   & 52246  & 14783   & 86187  & 116001 & 105581   & 10798  & 22381 & 6039   & 19448  & 9186  & 3523   \\ 
\# Validation & 80 & 68 & 45 & 33 & 67 & 107   & 126 & 68 & 38 & 8  & 8 & 1 & 1  \\ \midrule

VGG-16 (R) & \textbf{0.797} & \textbf{0.809} & 0.876  & 0.873 & 0.905 & 0.867 & 0.889  & 0.604& 0.877  & \textbf{0.807}  & 0.600 & \textbf{0.369} & 0.884 & 0.83 \\ 
VGG-16 (O)       & 0.764  & 0.699  & 0.783  & 0.811  & 0.802  & 0.767 & 0.836  & 0.552   & 0.827  & 0.459  & 0.587  & 0.172 & 0.828 & 0.76 \\ 
VGG-16 (F) & 0.776& 0.807 & \textbf{0.904}  & \textbf{0.931}  & \textbf{0.919}& \textbf{0.940} & \textbf{0.914} & \textbf{0.656}  & \textbf{0.880}  & 0.696  & \textbf{0.678} & 0.189 & \textbf{1.000} & \textbf{0.86} \\ \midrule

VGG-19 (R) & 0.802 & \textbf{0.824} & 0.852  & \textbf{0.913} & 0.913 & 0.754 & 0.882  & \textbf{0.767} & 0.867 & \textbf{0.786}  & \textbf{0.675} & 0.150 & 0.893&0.84  \\ 
VGG-19 (O)  & 0.747  & 0.734  & 0.815  & 0.799  & 0.808 & 0.765  & 0.812 & 0.534 & 0.834  & 0.523 & 0.534  & 0.326 & 0.867& 0.76 \\ 
VGG-19 (F)& \textbf{0.818} & 0.799 & \textbf{0.867}  & 0.912  & \textbf{0.918} & \textbf{0.916} & \textbf{0.903} & 0.677 & \textbf{0.872} & 0.618 & 0.627& \textbf{0.408}& \textbf{0.979} & \textbf{0.85}\\ \midrule

Inception (R) & 0.559 & 0.507 & 0.560  & 0.619 & 0.590 & 0.538 & 0.550 & 0.533 & 0.460 & 0.553 & 0.421 & 0.225 & 0.620 &0.54 \\ 
Inception (O) & 0.711 & 0.698 & 0.805  & 0.782 & 0.823 & 0.720 & 0.844 & 0.432 & 0.843 & 0.464  & 0.644 & 0.429 & 0.867 &0.74 \\ 
Inception (F) & \textbf{0.787} & \textbf{0.847} & \textbf{0.902} & \textbf{0.911} & \textbf{0.922} & \textbf{0.951} & \textbf{0.916}   & \textbf{0.639}  & \textbf{0.889} & \textbf{0.654}  & \textbf{0.810}   & \textbf{0.841} & \textbf{0.884} & \textbf{0.86}   \\ \midrule

Xception (R)& 0.606 & 0.652& 0.736  & 0.651 & 0.638  & 0.574  & 0.628 & 0.540 & 0.654 & 0.507& 0.537 & 0.129 & 0.931 & 0.63 \\ 
Xception (O) & 0.740 & 0.702 & 0.817  & 0.824 & 0.817 & 0.769 & 0.810 & 0.524 & 0.845  & 0.623 & 0.594 & 0.356  & 0.687 & 0.76 \\ 
Xception (F) & \textbf{0.833}  & \textbf{0.816}   & \textbf{0.905}  & \textbf{0.918}   & \textbf{0.933}     & \textbf{0.951}   & \textbf{0.909}  & \textbf{0.681}     & \textbf{0.868} & \textbf{0.684} & \textbf{0.757} & \textbf{0.923} & \textbf{0.974} & \textbf{0.87}  \\ \midrule

DenseNet-121(R) & \textbf{0.839} & 0.791   & 0.837  & 0.897   & 0.886    & 0.718  & 0.869 & \textbf{0.736} & 0.873 & 0.691& 0.722 & 0.004  & 0.858 & 0.83\\ 
DenseNet-121(O)& 0.717& 0.688 & 0.819  & 0.846 & 0.814 & 0.781 & 0.819    & 0.668  & 0.792   & 0.642 & 0.601  & 0.039   & 0.708  & 0.77\\ 
DenseNet-121(F)& 0.807 & \textbf{0.800} & \textbf{0.917}  & \textbf{0.924}  & \textbf{0.924} & \textbf{0.934}   & \textbf{0.916}  & 0.714   & \textbf{0.896} & \textbf{0.723}   & \textbf{0.777}  & \textbf{0.584}    & \textbf{0.918} & \textbf{0.87}\\ \midrule

ResNet-18 (R) & 0.815       & \textbf{0.847}   & 0.867  & 0.904 & 0.903 & 0.838 & 0.889 & \textbf{0.690}       & \textbf{0.917} & \textbf{0.796} & 0.661  & 0.215  & 0.957 &0.85 \\ 
ResNet-18 (O) & 0.745 & 0.721& 0.825  & 0.820    & 0.808 & 0.741  & 0.833  & 0.598  & 0.862   & 0.528  & 0.726 & 0.043  & 0.777 & 0.77\\ 
ResNet-18 (F) & \textbf{0.819}  & 0.820 & \textbf{0.892}  & \textbf{0.933}  & \textbf{0.939}   & \textbf{0.947} & \textbf{0.923}   & 0.655  & 0.876 & 0.715 & \textbf{0.838}   & \textbf{0.305} & \textbf{0.983} & \textbf{0.87}\\ \midrule

ResNet-50 (R) & 0.783 & 0.792 & 0.851  & 0.897  & 0.897 & 0.731 & 0.887  & 0.718 & \textbf{0.905} & \textbf{0.677}  & \textbf{0.808}  & 0.064 & 0.910 & 0.83\\ 
ResNet-50 (O) & 0.723 & 0.755 & 0.799  & 0.872 & 0.841 & 0.781 & 0.851  & \textbf{0.579}  & 0.843  & 0.579   & 0.726  & 0.069  & 0.811 & 0.78\\ 
ResNet-50 (F) & \textbf{0.792} & \textbf{0.848} & \textbf{0.901} & \textbf{0.924} & \textbf{0.904} & \textbf{0.930} & \textbf{0.912} & 0.577 & 0.892 & 0.642 & 0.774    & \textbf{0.837}         & \textbf{0.961}  & \textbf{0.85}       \\ \midrule

Ensemble (Avg.)  & 0.8254 & 0.8002 & 0.9121 & 0.9465   & 0.9316 & 0.9436  & 0.9194  & 0.5955 & 0.8827 & 0.6455 & 0.8008 & 0.2489  & 0.8584 & 0.86\\ 
Ensemble (Wted.) & 0.8262  & 0.8004 & 0.9122 & 0.9466 & 0.9318 & 0.9437 & 0.9197 & 0.6031  & 0.8829 & 0.6665 & 0.8070 & 0.4592 & 0.8841  & 0.86\\ \midrule

\cite{chexpert} & 0.811 & 0.840 & 0.929  & 0.932 & 0.931   & --  & --  & --  & -- & -- & -- & -- & -- \\ \bottomrule
\end{tabular}
}
\end{table*}

\renewcommand{\arraystretch}{1}
\begin{table*}[!tbh]
\centering
\label{Tab:u_ones}
\caption{Area under the ROC curves for the Chest X-ray disease classification using  U-Ones setting to handle uncertain labels in the training set. The best performance of each architecture is indicated in bold for each disease.}
\resizebox{\textwidth}{!}{
\begin{tabular}{@{}lccccccccccccc@{}}
\toprule
                & Atelectasis & \begin{tabular}[c]{@{}l@{}}Cardio-\\ megaly\end{tabular} & Edema & \begin{tabular}[c]{@{}l@{}}Consolid-\\ ation\end{tabular} & \begin{tabular}[c]{@{}l@{}}Pleural \\ Effusion\end{tabular} & \begin{tabular}[c]{@{}l@{}}Support \\ Devices\end{tabular} & \begin{tabular}[c]{@{}l@{}}Lung\\ Opacity\end{tabular} & \begin{tabular}[c]{@{}l@{}}Enlarged\\ Cardiom.\end{tabular} & \begin{tabular}[c]{@{}l@{}}No \\ Finding\end{tabular} & \begin{tabular}[c]{@{}l@{}}Pneum-\\ onia\end{tabular} & \begin{tabular}[c]{@{}l@{}}Pneum-\\ othorax\end{tabular} & \begin{tabular}[c]{@{}l@{}}Lung\\ Lesion\end{tabular} & \begin{tabular}[c]{@{}l@{}}Pleural \\ Other\end{tabular} \\ \midrule

VGG-16 (R)      & 0.831   & \textbf{0.812}   & \textbf{0.901} & 0.899    & \textbf{0.927}  & \textbf{0.853}      & \textbf{0.885}    & \textbf{0.585}    & 0.889     & 0.791   & \textbf{0.690}      & 0.185   & 0.936    \\ 
VGG-16 (O)      & 0.749   & 0.647  & 0.828 & 0.851    & 0.814   & 0.790                                                      & 0.823  & 0.524   & 0.849    & 0.632  & 0.511    & 0.124   & 0.755   \\ 
VGG-16 (F)      & \textbf{0.858}  & 0.756   & 0.870 & \textbf{0.915}   & 0.913    & 0.823        & 0.882    & 0.504                & \textbf{0.890}   & \textbf{0.824}      & 0.647   & \textbf{0.670}  & \textbf{0.966} \\ \midrule

VGG-19 (R)      & 0.857     & 0.792 & 0.888 & \textbf{0.901}  & 0.904  & 0.736    & 0.881  & 0.609   & 0.870  & \textbf{0.658}  & 0.673  & \textbf{0.309} & 0.747\\ 
VGG-19 (O)  & 0.767 & 0.772  & 0.831 & 0.795 & 0.786  & 0.777   & 0.844  & 0.503 & 0.861 & 0.655  & \textbf{0.790}  & 0.124 & \textbf{0.790}                                                   \\ 
VGG-19 (F)  & \textbf{0.873} & \textbf{0.821}   & \textbf{0.904} & 0.898  & \textbf{0.930}  & \textbf{0.918}  & \textbf{0.894}     & \textbf{0.721} & \textbf{0.903}  & 0.632  & 0.684   & 0.142 & 0.730 \\ \midrule

Inception (R)   & 0.508       & 0.514         & 0.508 & 0.500         & 0.500      & 0.500                               & 0.500     & \textbf{0.498}  & 0.526     & 0.585   & 0.480    & 0.755    & 0.659  \\ 
Inception (O)   & 0.757  & 0.682    & 0.815 & 0.810   & 0.836  & 0.778  & 0.854   & 0.420    & 0.828   & 0.715 & 0.656  & 0.193  & 0.695   \\ 
Inception (F)   & \textbf{0.849}  & \textbf{0.799}  & \textbf{0.916} & \textbf{0.863}   & \textbf{0.934}  & \textbf{0.939}   & \textbf{0.924}   & 0.411  & \textbf{0.881 }            & \textbf{0.830} & \textbf{0.816}  & \textbf{0.845}  &\textbf{ 0.867 }     \\ \midrule

Xception (R)    & 0.431  & 0.559       & 0.586 & 0.570         & 0.608      & 0.493      & 0.499     & 0.602                             & 0.410        & 0.606     & 0.514      & 0.339     & 0.313      \\ 
Xception (O)    & 0.759       & 0.700  & 0.852 & 0.839    & 0.797   & 0.756                                                      & 0.804   & 0.465  & 0.819    & \textbf{0.871}    & 0.653 & 0.189   & 0.773   \\ 
Xception (F)    & \textbf{0.833}       & \textbf{0.837} & \textbf{0.929} & \textbf{0.865}    & \textbf{0.949} & \textbf{0.948}   & \textbf{0.902}   & \textbf{0.655} & \textbf{0.853}  & 0.753 & \textbf{0.689}  & \textbf{0.506}  & \textbf{0.833}  \\ \midrule

DenseNet-121(R) & \textbf{0.827}  & 0.786    & 0.850 & \textbf{0.889}           & 0.890        & 0.726          & 0.866       & \textbf{0.713} & \textbf{0.877}     & 0.799        & 0.715                                                    & 0.017    & 0.880                                                    \\ 
DenseNet-121(O) & 0.772   & 0.667                                                    & 0.866 & 0.842                                                     & 0.837  & 0.775    & 0.833  & 0.497  & 0.825   & 0.777  & 0.703   & 0.163  & 0.721 \\ 
DenseNet-121(F) & 0.818   & \textbf{0.818}  & \textbf{0.923} & 0.860   & \textbf{0.938}  & \textbf{0.927}  & \textbf{0.921} & 0.466    & 0.875 & \textbf{0.811}  & \textbf{0.765} & \textbf{0.403}  & \textbf{0.901} \\ \midrule

ResNet-18 (R)   & \textbf{0.857}  & 0.825  & 0.873 & \textbf{0.916}   & 0.906   & 0.803  & 0.888 & \textbf{0.724}   & \textbf{0.904}   & \textbf{0.850}       & 0.688  & 0.094   & 0.966 \\ 
ResNet-18 (O)   & 0.752   & 0.737  & 0.853 & 0.837  & 0.785 & 0.729    & 0.801   & 0.510  & 0.857  & 0.642               & 0.659  & 0.060   & 0.464  \\ 
ResNet-18 (F)   & 0.821    & \textbf{0.836}  & \textbf{0.894} & 0.908    & \textbf{0.925}   & \textbf{0.926}   & \textbf{0.904}  & 0.616                               & 0.903  & 0.757   & \textbf{0.801} & \textbf{0.476}   & \textbf{0.974}  \\ \midrule

ResNet-50 (R)   & 0.807  & 0.781    & 0.855 & 0.900   & 0.877   & 0.683  & 0.861    & \textbf{0.645}                              & 0.876   & 0.773   & 0.811     & 0.146    & 0.730   \\ 
ResNet-50 (O)   & 0.785   & 0.713    & 0.829 & 0.885  & 0.842   & 0.774   & 0.836    & 0.560   & 0.856  & \textbf{0.809}   & 0.707   & 0.120  & 0.631   \\ 
ResNet-50 (F)   & \textbf{0.859}    & \textbf{0.828} & \textbf{0.910} & \textbf{0.901}   & \textbf{0.924}  & \textbf{0.918}     & \textbf{0.912}  & 0.550  & \textbf{0.898}  & 0.801 & \textbf{0.826}  & \textbf{0.592} & \textbf{0.871}  \\ \bottomrule
\end{tabular}
}
\end{table*}

\renewcommand{\arraystretch}{1}
\begin{table*}[!tbh]
\centering
\label{Tab:u_ignore}
\caption{Area under the ROC curves for the Chest X-ray disease classification using  U-Ignore setting to handle uncertain labels in the training set. The best performance of each architecture is indicated in bold for each disease.}
\resizebox{\textwidth}{!}{
\begin{tabular}{@{}lccccccccccccc@{}}
\toprule
                  & Atelectasis & \begin{tabular}[c]{@{}l@{}}Cardio-\\ megaly\end{tabular} & Edema & \begin{tabular}[c]{@{}l@{}}Consolid-\\ ation\end{tabular} & \begin{tabular}[c]{@{}l@{}}Pleural \\ Effusion\end{tabular} & \begin{tabular}[c]{@{}l@{}}Support \\ Devices\end{tabular} & \begin{tabular}[c]{@{}l@{}}Lung\\ Opacity\end{tabular} & \begin{tabular}[c]{@{}l@{}}Enlarged\\ Cardiom.\end{tabular} & \begin{tabular}[c]{@{}l@{}}No \\ Finding\end{tabular} & \begin{tabular}[c]{@{}l@{}}Pneum-\\ onia\end{tabular} & \begin{tabular}[c]{@{}l@{}}Pneum-\\ othorax\end{tabular} & \begin{tabular}[c]{@{}l@{}}Lung\\ Lesion\end{tabular} & \begin{tabular}[c]{@{}l@{}}Pleural \\ Other\end{tabular} \\ \midrule
VGG-16 (R)      & \textbf{0.844}       & \textbf{0.844}        & 0.876 & \textbf{0.922}         & \textbf{0.940}            & 0.872           & 0.892        & \textbf{0.650}             & \textbf{0.876}      & \textbf{0.906}     & 0.645        & 0.142       & 0.957         \\ 
VGG-16 (O)      & 0.746       & 0.684        & 0.772 & 0.742         & 0.810            & 0.751           & 0.814        & 0.597             & 0.832      & 0.447     & 0.544        & \textbf{0.206}       & 0.631         \\ 
VGG-16 (F)      & 0.819       & 0.773        & \textbf{0.902} & 0.918         & 0.923            & \textbf{0.937}           & \textbf{0.898}        & 0.563             & 0.871      & 0.709     & \textbf{0.804}        & 0.116       & \textbf{0.966}         \\ \midrule

VGG-19 (R)      & \textbf{0.819}       & \textbf{0.824}        & 0.859 & 0.916         & 0.922            & 0.753           & 0.879        & \textbf{0.758}             & \textbf{0.887}      & 0.626     & 0.631        & 0.176       & \textbf{0.948}         \\ 
VGG-19 (O)      & 0.739       & 0.745        & 0.839 & 0.807         & 0.822            & 0.803           & 0.839        & 0.561             & 0.848      & 0.531     & 0.619        & 0.292       & 0.725         \\ 
VGG-19 (F)  & 0.811 & 0.790 & \textbf{0.912} &  \textbf{0.918} & \textbf{0.929} & \textbf{0.949} & \textbf{0.911} & 0.658 & 0.845 & \textbf{0.685} & \textbf{0.732} & \textbf{0.639}  &  0.936 \\ \midrule

Inception (R)   & 0.457       & 0.527        & 0.484 & 0.557         & 0.500            & 0.523           & 0.487        & \textbf{0.523}             & 0.436      & 0.596     & 0.611        & \textbf{0.646}       & 0.826         \\ 
Inception (O)   & 0.738       & 0.717        & 0.804 & 0.788         & 0.820            & 0.750            & 0.830        & 0.439             & 0.824      & 0.565     & 0.566        & 0.292       & 0.335         \\ 
Inception (F)   & \textbf{0.864}       & \textbf{0.824}        & \textbf{0.907} & \textbf{0.941}         & \textbf{0.926}            & \textbf{0.944}           & \textbf{0.929}        & 0.509             & \textbf{0.885}      & \textbf{0.767}     & \textbf{0.881}        & 0.232       & \textbf{0.961}         \\ \midrule

Xception (R)    & 0.527       & 0.646        & 0.712 & 0.653         & 0.647            & 0.536           & 0.622        & 0.473             & 0.631      & 0.494     & 0.629        & \textbf{0.391}       & 0.378         \\ 
Xception (O)    & 0.731       & 0.725        & 0.819 & 0.831         & 0.817            & 0.742           & 0.827        & 0.532             & 0.840      & 0.711     & 0.618        & 0.009       & 0.768         \\ 
Xception (F)    & \textbf{0.813}       & \textbf{0.817}        & \textbf{0.906} & \textbf{0.946}         & \textbf{0.923}            & \textbf{0.950}            & \textbf{0.919 }       & \textbf{0.592}             & \textbf{0.892 }     &\textbf{ 0.742 }    & \textbf{0.846}        & 0.335       & \textbf{0.983 }        \\ \midrule

DenseNet-121(R) & \textbf{0.850}       & 0.793        & 0.863 & 0.914         & 0.891            & 0.698           & 0.881        & \textbf{0.759 }            & 0.899      & 0.694     & 0.727        & 0.026       & 0.914         \\ 
DenseNet-121(O) & 0.741       & 0.725        & 0.841 & 0.855         & 0.842            & 0.768           & 0.855        & 0.587             & 0.854      & 0.629     & 0.582        & 0.193       & 0.725         \\ 
DenseNet-121(F) & 0.821       & \textbf{0.815}        & \textbf{0.901} & \textbf{0.927}         & \textbf{0.945}            & \textbf{0.945}           & \textbf{0.919}        & 0.630             & \textbf{0.905}      & \textbf{0.696}     & \textbf{0.817}        & \textbf{0.528}       & \textbf{0.970}         \\ \midrule

ResNet-18 (R)   & \textbf{0.841}       & \textbf{0.824}        & 0.874 & \textbf{0.921}         & \textbf{0.923}            & \textbf{0.831}           & \textbf{0.886}        & \textbf{0.700}             & \textbf{0.886}      & \textbf{0.787}     & 0.748        & 0.112       & \textbf{0.970}         \\ 
ResNet-18 (O)   & 0.713       & 0.725        & 0.846 & 0.801         & 0.795            & 0.724           & 0.806        & 0.592             & 0.872      & 0.527     & \textbf{0.766}        & 0.069       & 0.914         \\ 
ResNet-18 (F)   & 0.771       & 0.792        & \textbf{0.876} & 0.903         & 0.908            & \textbf{0.831}           & 0.868        & 0.623             & 0.874      & 0.757     & 0.726        & \textbf{0.472}       & 0.957         \\ \midrule

ResNet-50 (R)   & \textbf{0.843}       & 0.799        & 0.859 & 0.914         & 0.900            & 0.716           & 0.887        & \textbf{0.680}             & \textbf{0.902}      & \textbf{0.748}     & 0.768        & 0.112       & 0.721         \\ 
ResNet-50 (O)   & 0.769       & 0.734        & 0.811 & 0.894         & 0.842            & 0.772           & 0.836        & 0.573             & 0.864      & 0.658     & 0.692        & 0.047       & 0.803         \\ 
ResNet-50 (F)   & 0.839       & \textbf{0.856}        & \textbf{0.910}  & \textbf{0.927}         & \textbf{0.928}            & \textbf{0.952}           & \textbf{0.915}        & 0.643             & 0.892      & 0.743     & \textbf{0.834}        & \textbf{0.167}       & \textbf{0.961}         \\ \bottomrule
\end{tabular}
}
\end{table*}

\renewcommand{\arraystretch}{1.1}
\begin{table*}[h]
\centering
    \caption{Compute time complexity of the experiments during training. Reported values are the training time (minutes) on a 11.3 FP32 TensorTFLOPS Nvidia GTX 1080Ti for forward and backward operation on the network on a batch size of 16. The last row sums up the total time per architecture across all explorations performed. The training time for the Off the shelf setting can be further reduced by precomputing the final features of the CNN to reduce redundant computations in the forward and backward pass but this was not implemented in our setup. 
    }
    \label{tab:computetime}
    \resizebox{\textwidth}{!}{

\begin{tabular}{@{}llccccccc@{}}
\toprule
Weight initialization & Uncertain Labels   & VGG-16 & VGG-19 & ResNet-18 & ResNet-50 & Inception & Xception & DenseNet-121 \\ \midrule
Random (R)         & U-Zeros                      & 464    & 518    & 275       & 550       & 256       & 442      & 348          \\ 
                                   & U-Ones                       & 462    & 524    & 188       & 493       & 262       & 437      & 344          \\ 
                                   & U-Ignore                     & 463    & 597    & 199       & 458       & 268       & 449      & 346          \\ 
Off the shelf (O)  & U-Zeros                      & 461    & 479    & 198       & 195       & 201       & 300      & 324          \\ 
                                   & U-Ones                       & 459    & 474    & 198       & 200       & 206       & 349      & 318          \\ 
                                   & U-Ignore                     & 462    & 485    & 209       & 350       & 209       & 296      & 312          \\ 
Fine tuned (F)      & U-Zeros                      & 464    & 519    & 204       & 494       & 494       & 453      & 345          \\ 
                                   & U-Ones                       & 464    & 524    & 276       & 548       & 547       & 454      & 333          \\ 
                                   & U-Ignore                     & 456    & 514    & 240       & 506       & 503       & 456      & 398          \\ \midrule
Total time (mins)                  & \begin{tabular}[c]{@{}l@{}}6 epochs; \\ 13,963 batch/epoch\end{tabular} &   4155     &  4634      & 1987      &  3794         &    2946       &  3636        &      3068        \\ \bottomrule
\end{tabular}
}
\end{table*}

\end{document}